\documentclass[runningheads]{llncs}
\usepackage{graphicx}
\usepackage{graphicx}
\usepackage{caption}
\usepackage{subcaption}
\usepackage{amsmath}
\begin{document}
\title{On Additive Gaussian Processes for Wind Farm Power Prediction}
%
%
\author{Simon M.\ Brealy\inst{1} \and
Lawrence A.\ Bull\inst{2} \and
Daniel S.\ Brennan\inst{1} \and
Pauline Beltrando\inst{3} \and
Anders Sommer\inst{3} \and
Nikolaos Dervilis\inst{1} \and
Keith Worden\inst{1}}
%
\authorrunning{S. Brealy et al.}
%
\institute{Dynamics Research Group, Department of Mechanical Engineering, The University of Sheffield, Mappin Street, Sheffield, S1 3JD, UK
\email{ssmith8@sheffield.ac.uk}\\
\and
Computational Statistics and Machine Learning Group, Department of Engineering, University of Cambridge, Cambridge, CB3 0FA, UK
\and
Vattenfall R\&D, Vattenfall AB, Älvkarleby, Sweden, \email{pauline.beltrando@vattenfall.com}
}
\maketitle              
\begin{abstract}

Population-based Structural Health Monitoring (PBSHM) aims to share information between similar machines or structures. %
This paper takes a population-level perspective, exploring the use of additive Gaussian processes to reveal variations in turbine-specific and farm-level power models over a collected wind farm dataset. %
The predictions illustrate patterns in wind farm power generation, which follow intuition and should enable more informed control and decision-making.

\end{abstract}
\section{Introduction}
Population-based Structural Health Monitoring (PBSHM) was developed to address some of the challenges of SHM, which include: (1) a lack of labelled damage-state data, (2) short or incomplete histories of failure data, and (3) difficulties resulting from operational and environmental variability. %
To do this, PBSHM aims to utilise information from across a variety of structures to perform inferences that generalise well for the entire population~\cite{gardner_2020}. %
Three wind farm datasets have recently become available for research, spanning a total of 224 turbines.



\subsection{An application: Wind power forecasting}
An area of research that has received considerable attention, which could also benefit from population-based approaches, is wind power forecasting. %
With the increasing proportion of power generation from weather-dependent renewable energy sources, the additional uncertainty they bring makes the task of balancing supply and demand in electricity networks more difficult. %
To help manage this, predictions of future solar and wind generation are used for risk mitigation in both operational and electricity market settings; this includes planning unit dispatch, maintenance scheduling and maximising profit in power trading~\cite{Gilbert_2021,hanifi_2020}. 



In recent times, probabilistic approaches to power forecasting have gained greater attention from researchers~\cite{Bazionis_2021}---which, unlike deterministic approaches, provide information on the expected variance of the prediction, giving insight into uncertainty which could be helpful in the risk-mitigation process.

Wind turbines are known to produce wakes, resulting in spatio-temporal variations in the wind field across wind farms~\cite{lin_2023}. %
Given that the SCADA data consists of measurements across this wind field over time, effective utilisation of this data in power forecasting at the wind farm and fleet levels may lead to improved predictive accuracy for risk mitigation. 

\subsection{Contribution}

In Section~\ref{section:eda}, the data from the wind farm Supervisory Control and Data Acquisition Systems (SCADA) are explored and pre-processed to be made suitable for wind power forecasting. As a first step, additive Gaussian process models (described in Section~\ref{section:agp}) were fitted to the data, in order to predict power. This form of model was chosen as it is both highly flexible and interpretable. Results and Conclusions are discussed in Sections~\ref{section:results} and~\ref{section:conclusions} respectively. Finally, it is to be noted that all data shown in plots has been normalised and/or transformed to preserve anonymity to the data provider.

\section{Data Exploration}\label{section:eda}
The SCADA data covers 224 wind turbines, spanning three separate wind farms between the years 2020-2023, at a ten-minute resolution. The measurements cover a broad range of areas, but generally fall into electrical, control, component/system temperature measurements, and ambient environment parameters. 

For this initial study---and the motivation of wind power forecasting, only data from within the year of 2021 for one of the wind farms (identified as \textit{Ciabatta} here) was considered; this was chosen to reduce computational overhead, maintain a full year of cyclical environmental variation, and to demonstrate the modelling approach.

\begin{figure}[t]
\centering
\begin{subfigure}{.5\textwidth}
  \centering
  \includegraphics[width=\textwidth]{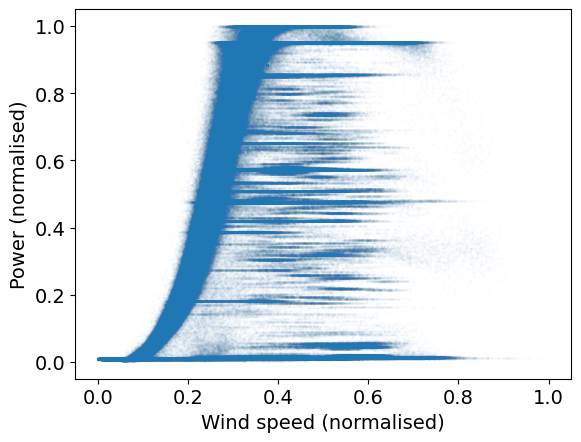}
  \caption{Individual turbines.}
  \label{fig:1sub1}
\end{subfigure}%
\begin{subfigure}{.5\textwidth}
  \centering
  \includegraphics[width=\textwidth]{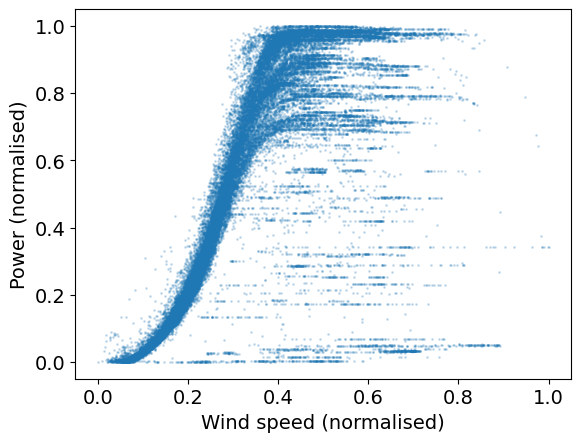}
  \caption{Aggregate wind farm.}
  \label{fig:1sub2}
\end{subfigure}
\caption{Normalised data showing the power curve relationships for wind farm \textit{Ciabatta}. Plot (a) shows data at the individual turbine level, whilst plot (b) shows data at the aggregate wind farm level.}
\label{fig:1}
\end{figure}

The raw unfiltered data showing the power-curve relationship at individual wind turbines are shown in Fig.~\ref{fig:1sub1}. Fig.~\ref{fig:1sub2} shows the same data, but aggregated (i.e.\ summed) at the wind farm level. %
Median values were taken for the wind speed to minimise the effect of outliers, whilst power was summed across all turbines for each timestamp.

A number of patterns typical to wind farm operation are identifiable. %
These include: curtailments (whereby turbine power is limited), shutdowns (indicated by a power output of zero, despite significant wind speeds), and power boosting (where power is raised above nominal turbine capacity). %
Given that these trends represent controlled deviations away from the nominal power curve, these data were filtered as described in Section~\ref{section:preprocessing} such that the forecast predicts the total \textit{available} power.

\subsection{Preprocessing}\label{section:preprocessing}
A range of filters were implemented to remove controlled deviations from the nominal power curve; these were based on knowledge of turbine properties such as cut-in and rated wind speed and power, as well as turbine control parameters. 
Following the initial filtering steps, an additional filter based on the Mahalanobis squared-distance metric was used~\cite{worden_2000}. %
This filter was applied to the blade pitch angle vs. power relationship, as visually it was found to better separate the remaining outliers. %

\begin{figure}[t]
\centering
\begin{subfigure}{.5\textwidth}
  \centering
  \includegraphics[width=\textwidth]{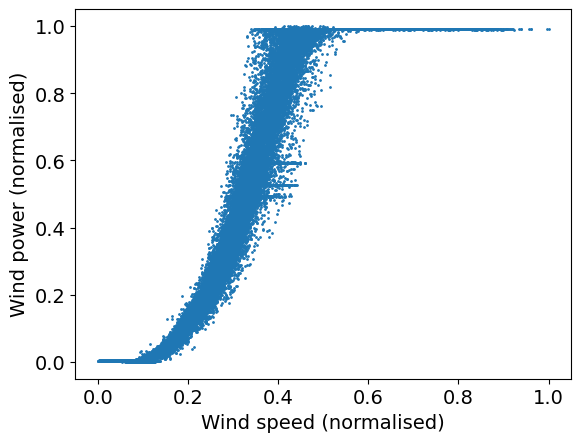}
  \caption{Individual turbines.}
  \label{fig:2sub1}
\end{subfigure}%
\begin{subfigure}{.5\textwidth}
  \centering
  \includegraphics[width=\textwidth]{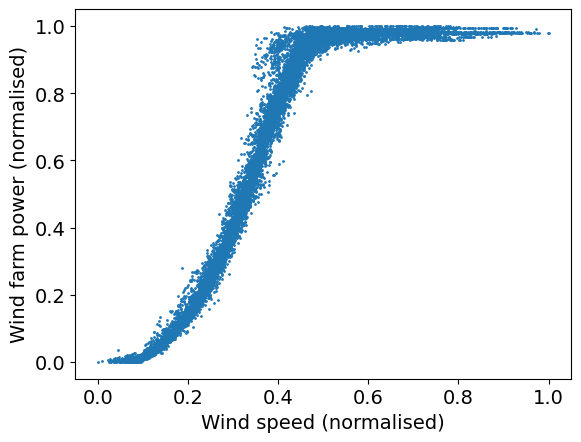}
  \caption{Aggregate wind farm.}
  \label{fig:2sub2}
\end{subfigure}
\caption{Filtered data showing the power curves for both individual turbines (a), and at the aggregate wind farm level (b)}
\label{fig:2}
\end{figure}

Fig.~\ref{fig:2} shows the filtered versions of the power curve data, where plot (a) shows the individual turbine level, and (b) shows the wind farm aggregate level. In~\ref{fig:2}(a), a small region of possible curtailments remain between a normalised power of 0.5 - 0.6, whilst some small curtailments for individual turbines in (b) have likely resulted in the larger variance seen during the rated power region of the curve. However, from a visual perspective, the data is now much more representative of the normal power curve.

\begin{figure}
\centering
\begin{subfigure}{.5\textwidth}
  \centering
  \includegraphics[width=\textwidth]{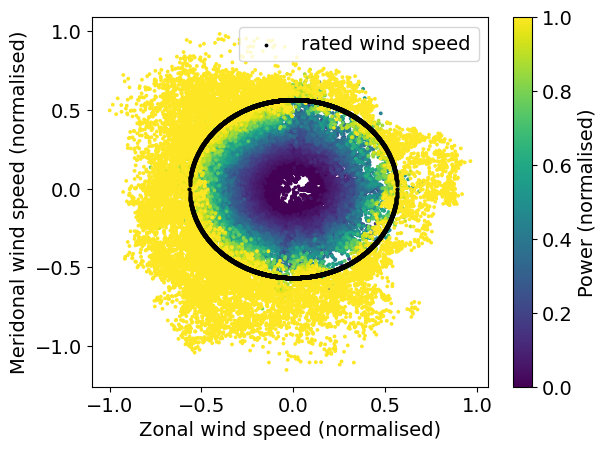}
  \caption{Turbine \textit{X} located on the western side.}
  \label{fig:3sub1}
\end{subfigure}%
\begin{subfigure}{.5\textwidth}
  \centering
  \includegraphics[width=\textwidth]{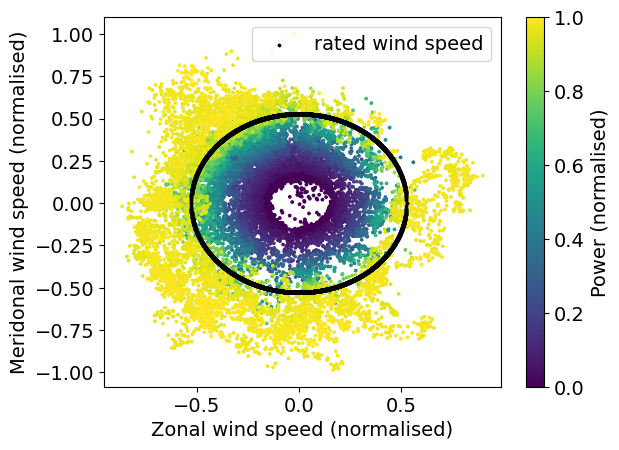}
  \caption{Aggregate wind farm.}
  \label{fig:3sub2}
\end{subfigure}
\caption{Filtered data showing the zonal and meridional wind speeds, coloured by power output for both an turbine \textit{X} (a), and at the aggregate wind farm level~(b)}
\label{fig:3}
\end{figure}

To demonstrate the wake effect, Fig.~\ref{fig:3} shows the effect of wind speed and direction on the output power, at both an individual turbine (a) (referred to as turbine \textit{X}), and at the aggregate farm level (b). Here, wind direction is determined using the turbine yaw angle as a proxy, which, in turn, is used to split the free stream wind speed (wind farm maximum) into zonal and meridional components. %
Data are coloured by power output, whilst the black circles represent turbine-rated power as a reference point. 

In~\ref{fig:3}(a), turbine \textit{X} was deliberately chosen as it is located on the edge of the wind farm; as such, it is not expected to be significantly influenced by wakes when wind is coming from the western direction (in this case). While subtle, this can be seen in the data, where winds with a westerly component (between the six and 12 o'clock position) generally appear to result in more consistent and perhaps higher power outputs, in contrast to winds with an easterly component. At the aggregate level in~\ref{fig:3}(b), as might be expected, the directional relationship of wind versus power appears more complex. Capturing this complexity may require a relatively flexible model.

\section{Modelling Approach}\label{section:agp}

\subsection{Generalised Additive Models}
Generalised Additive Models (GAMs) were first introduced in~\cite{Hastie_1986,Hastie_1990}, and are a generalisation of many popular statistical models, such as linear regression, logistic regression, and Generalised Linear Regression. GAMs have the following additive structure, 

\begin{equation}
f(\mathbf{x})=g\left(f_1\left(x_1\right)+f_2\left(x_2\right)+\cdots+f_D\left(x_D\right)\right) 
\end{equation}

\noindent where the output $f(\mathbf{x})$ is modelled as a sum of individual predictor functions $\{f_1 : f_D\}$ transformed by a link function, $g(\cdot)$. %
In this work, the individual functions are smooth and regularised (helping to avoid overfitting), non-parametric (requiring no prior assumption on their forms) and can capture a non-linear, non-stationary response. %
GAM architectures are also interpretable, and in this application, the relationships of individual variables $\{x_1 : x_D\}$ can be easily inspected.

\subsection{Gaussian Processes}
Gaussian Processes (GPs) can be used for classification or regression~\cite{rasmussen_2006} and are commonly described as being a distribution over functions. %
This distribution function can then be updated using Bayesian inference after having observed values from measurement. %
Rather than isolating subsets of variables (as is the case for GAMs), in standard GP regression, the model typically depends on all variables simultaneously, with the following structure,
\begin{align}
    \mathbf{y} &= f\left(\mathbf{x}\right) + \mathbf{\epsilon}
\end{align}

\noindent where the output $\mathbf{y}$ is sampled from $f(\mathbf{x})$ operating on a D-dimensional input $\mathbf{x}$, with additive noise $\epsilon$. In a Bayesian manner, prior distributions are placed over the latent functions and variables,

\begin{align}
    \mathbf{\epsilon} &\sim \mathcal{N}\left(0, \sigma_n^2\right) \\
    f &\sim \textrm{GP}(m(\mathbf{x}), k(\mathbf{x},\mathbf{x}'))
\end{align}

\noindent i.e. a Gaussian Process prior over the functions space, with a mean function $m$ and kernel function $k$, as well as additive noise that is normally distributed with variance $\sigma_n^2$. %

Since a zero-mean GP is assumed in this work ($m(\mathbf{x})=0$), the model structure is determined by its kernel $k\left(\mathbf{x}, \mathbf{x}'\right)$, which defines the prior covariance between any two inputs $\mathbf{x}$ and $\mathbf{x}'$. %
A common choice of the kernel function (which is used in the work thus far) is the squared exponential kernel,

\begin{equation}\label{eq:sq_exp}
    k\left(\mathbf{x}, \mathbf{x}'\right)=\sigma_f^2 \exp \left\{-\frac{1}{2 l^2}\mid\mid\mathbf{x}-\mathbf{x}'\mid\mid^2\right\}
\end{equation}

\noindent where $\sigma_f$ and $l$ are known as hyperparameters. $\sigma_f$ is the process variance, defining the variance of the expected functions about the mean, whilst the length scale, $l$, determines the rate at which the correlation between outputs decays across the input space (which can be thought of as function smoothness). For the interested reader, a more detailed explanation of Gaussian Processes and choices of kernel function is given here~\cite{rasmussen_2006}.

\subsection{Additive GPs}
A form of model that generalises and combines the benefits of both GAMs and GPs, is the additive Gaussian Process~\cite{Duvenaud_2011}. %
This is achieved via a kernel which allows additive interactions of all orders, ranging from first-order interactions (like in a GAM) all the way to \textit{D\textsuperscript{th}}-order interactions---as with the \textit{multivariate input} squared-exponential GP above. In a similar way to GAMs, additive GPs can be decomposed into independent and interpretable predictors, for subsets of input dimensions, which show their influence on the overall output.

Here, an additive kernel is used, with a separate process variance and length scale for each dimension,
\begin{equation}
k_{\textrm{add}}\left(\mathbf{x}, \mathbf{x}^{\prime}\right)= \sum_{i=1}^D k_i(x_i, x_i^{\prime} \mid \sigma{_f}{_i},l_i)
\end{equation}

\noindent This approach can easily be extended to higher orders using product kernels to combine dimensions. Given training data with inputs $\mathbf{X}$ ($N$ rows of $D$-dimensional vectors) and corresponding outputs $\mathbf{y}$ (vector length $N$) the GP predictive distributions can be decomposed into additive parts, with the mean prediction $\overline{f_i}(\cdot)$ defined as,


\begin{equation}
\overline{f_i}(x_i^{\star})=K_i\left(x_i^{\star}, \mathbf{X}[:, i]\right) \left(K_{\textrm{add}}+\sigma_n^2 I\right)^{-1} \mathbf{y}
\end{equation}

\noindent where $k_i(\cdot)$ is the kernel for the $i^{th}$ dimension, and $x_i^*$ are the points to be predicted for the $i^{th}$ dimension. %
A capital $K$ represents a kernel \textit{matrix}~\footnote{Note that $K$ becomes a vector if either input is a single observation.} where $K[n, m] = k_{\text {add}}(\mathbf{x}_n, \mathbf{x}_m)$. %
These components can also be used to determine the overall prediction,


\begin{equation}
\overline{f}(\mathbf{x}^\star)=K_{\textrm{add}}\left(\mathbf{x}^{\star}, \mathbf{X}\right) \left(K_{\textrm{add}}+\sigma_n^2 I\right)^{-1} \mathbf{y}
\end{equation}

\noindent where $\mathbf{x}^\star$ are the points to be predicted across all dimensions. For brevity, the associated equations for the posterior predictive variance are not included here, but can be found in~\cite{rasmussen_2006}.

\subsection{Method}


Three additive GP models were constructed to predict power output---one for turbine \textit{X} (located on the western edge), one for turbine \textit{Y} (located on the eastern edge) and one for the aggregate farm level. The turbine yaw angle (used as a proxy for wind direction) was split into two features of the sine and cosine of the angle; this was done to ensure a smooth transition between the equivalent yaw angles of 0 and 360 degrees. These two features, along with the freestream wind speed were used in first-order additive kernels. 

As the data here are not Gaussian distributed, an inverse sigmoid link function (for $g(\cdot)$) is used to transform the data into a space where it is approximately Gaussian and homoscedastic. %
The models were trained and tuned using a computationally-efficient \texttt{JAX} implementation~\cite{jax2018github}, on subsets of the filtered data. These subsets consisted of 5000 rows, determined using stratified sampling (based on the yaw angle) to help ensure consistent directional data coverage. Hyperparameters were tuned using a type II maximum likelihood approach, to minimise the negative log-marginal-likelihood for which there is an analytical solution~\cite{rasmussen_2006}. Predictions were made using a gridded method over the input domains.

\section{Results}\label{section:results}

\begin{figure}
\centering
\begin{subfigure}{.32\textwidth}
  \centering
  \includegraphics[width=\textwidth]{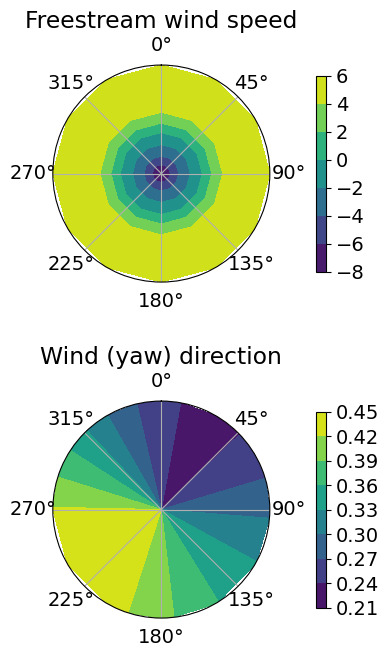}
  \caption{Turbine \textit{X}}
  \label{fig:4sub1}
\end{subfigure}%
\begin{subfigure}{.32\textwidth}
  \centering
  \includegraphics[width=\textwidth]{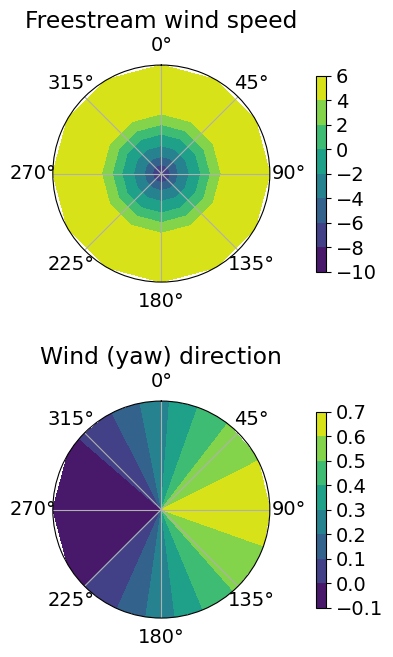}
  \caption{Turbine \textit{Y}}
  \label{fig:4sub2}
\end{subfigure}
\begin{subfigure}{.33\textwidth}
  \centering
  \includegraphics[width=\textwidth]{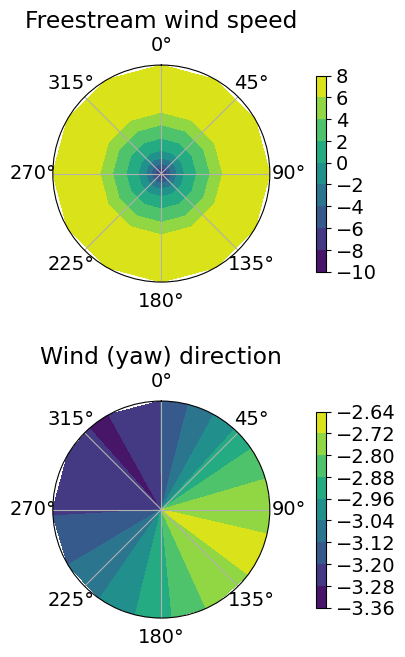}
  \caption{Aggregate wind farm}
  \label{fig:4sub3}
\end{subfigure}
\caption{Freestream wind and summed directional component mean predictions for turbine \textit{X} (\ref{fig:4sub1}), turbine \textit{Y} (\ref{fig:4sub2}), and wind farm \textit{Ciabbata} (\ref{fig:4sub3}).  The colours show the relative impact on the predicted power (in the transformed space)}
\label{fig:4}
\end{figure}

Fig~\ref{fig:4} shows the mean predictions from the additive GPs, for the freestream wind component, and the sine and cosine components summed to represent the wind directional component. Predictions further from the centre of each circle represent higher wind speeds, whilst the labelled angles represent the direction from which the wind is blowing. Plots are coloured by relative influence on the power prediction.

For turbine \textit{X} (located on the western side), the directional component is strongest when winds blow from the southwest. Conversely, for turbine \textit{Y} (located on the eastern side) the wind-directional component is strongest when winds blow from the east. These two observations are in line with intuition, as the turbines do not lie in the wake of other turbines in these circumstances.

At the aggregate farm level (Fig~\ref{fig:4sub3}), the directional component is strongest with winds from the southeast, and weakest from the northwest. %
Whilst the intuition is not as clear, on observation of the wind farm layout (not described here for anonymity), the result does not seem implausible. It is worth noting that in all cases, the effect of wind direction is comparatively small compared to the wind speed (as shown by the scale bars).

For all models, the predicted output power increases with wind speed monotonically and similarly in all directions; this is as expected, as the wind speed component should be insensitive to wind direction, since the wind direction component scales this out.

\section{Conclusions and next steps}\label{section:conclusions}

Additive Gaussian Process models were fitted to operational wind farm data, at both individual turbine and aggregate wind farm level to predict power output. The approach provides a highly-interpretable way of modelling the variations in power models within wind farms. In its current form, the model could be used as part of a probabilistic novelty detection scheme for performance and/or condition monitoring; it is likely however that there will be frequent false positives because of periods of curtailment.

The current approach could be refined further by: (1) improving the filtering method further to reduce curtailment noise, (2) adding additional explainable variables that may influence output power, (3) adding higher-order kernels to capture more complex relationships, or (4) using sparse-GP approximations to effectively increase the size of the training dataset.

By incorporating weather-forecast data, the additive GP can be extended to carry out probabilistic wind power forecasting at the individual turbine, farm, or even fleet levels, if appropriate aggregation is used. Furthermore, combining power forecasting with fatigue and foundation models could help move towards optimal decision-making in view of life cycle management and output control.


 



\section{Acknowledgements}
The authors gratefully acknowledge the support of the UK Engineering and Physical Sciences Research Council (EPSRC) and the Natural Environment Research Council (NERC), via grant references EP/W005816/1, EP/R003645/1 and EP/S023763/1. The authors also gratefully acknowledge Vattenfall R\&D for their time and for providing access to the data used in this study. For the purpose of open access, the author(s) has/have applied a Creative Commons Attribution (CC BY) licence to any Author Accepted Manuscript version arising.

%
%
%
\bibliographystyle{splncs04}
\bibliography{mybibliography}

@phdthesis{Gilbert_2021,
    author = {Ciaran Gilbert},
    title = {Topics in High Dimensional Energy Forecasting},
    school = {University of Strathclyde},
    year = 2021}

@inproceedings{Duvenaud_2011,
    author = {Duvenaud D.K., Nickisch H., Rasmussen C.},
    booktitle = {Advances in Neural Information Processing Systems},
    pages = {},
    publisher = {Curran Associates, Inc.},
    title = {{Additive} {Gaussian} {Processes}},
    volume = {24},
    year = {2011}
}

@book{Hastie_1990,
    author = {Hastie T., Tibshirani R.},
    title = {Generalized Additive Models},
    publisher = {Chapman and Hall},
    year = 1990
}

@article{Hastie_1986,
    author = {Hastie T., Tibshirani R.},
    title = {Generalized Additive Models},
    journal = {Statistical Science \textbf{1}, 297--318},
    year = 1986
}

@article{Bazionis_2021,
    author = {Bazionis I.K., Georgilakis P.S.},
    title = {Review of Deterministic and Probabilistic Wind Power Forecasting: Models, Methods, and Future Research},
    journal = {Electricity \textbf{2} 13--47},
    year = 2021 
}

@article{hanifi_2020,
	title = {A {Critical} {Review} of {Wind} {Power} {Forecasting} {Methods}—{Past}, {Present} and {Future}},
	volume = {13},
	copyright = {http://creativecommons.org/licenses/by/3.0/},
	issn = {1996-1073},
	language = {en},
	number = {15},
	urldate = {2023-07-20},
	journal = {Energies},
	author = {Hanifi, Shahram and Liu, Xiaolei and Lin, Zi and Lotfian, Saeid},
	year = {2020},
	pages = {3764},
}

@inproceedings{lin_2023,
	title = {A {Spatial} {Autoregressive} {Approach} for {Wake} {Field} {Prediction} {Across} a {Wind} {Farm}},
	language = {en},
	booktitle = {European {Workshop} on {Structural} {Health} {Monitoring} {EWSHM}},
	publisher = {Springer International Publishing},
	author = {Lin, Weijiang and Worden, Keith and Cross, Elizabeth},
	year = {2022},
	pages = {530--540},
}

@article{worden_2000,
	title = {Damage {Detection} {Using} {Outlier} {Analysis}},
	volume = {229},
	language = {en},
	number = {3},
	urldate = {2023-12-04},
	journal = {Journal of Sound and Vibration},
	author = {Worden, K. and Manson, G. and Fieller, N.R.J.},
	year = {2000},
	pages = {647--667},
}

@book{rasmussen_2006,
	address = {Cambridge},
	series = {Adaptive computation and machine learning},
	title = {Gaussian Processes for Machine Learning},
	isbn = {978-0-262-18253-9},
	language = {en},
	publisher = {MIT Press},
	author = {Rasmussen, Carl Edward and Williams, Christopher K. I.},
	year = {2006},
}

@software{jax2018github,
  author = {James Bradbury and Roy Frostig and Peter Hawkins and Matthew James Johnson and Chris Leary and Dougal Maclaurin and George Necula and Adam Paszke and Jake Vander{P}las and Skye Wanderman-{M}ilne and Qiao Zhang},
  title = {{JAX}: composable transformations of {P}ython+{N}um{P}y programs},
  url = {http://github.com/google/jax},
  version = {0.3.13},
  year = {2018},
}

@article{gardner_2020,
	title = {On the application of domain adaptation in structural health monitoring},
	volume = {138},
	issn = {0888-3270},
	language = {en},
	urldate = {2023-01-25},
	journal = {Mechanical Systems and Signal Processing},
	author = {Gardner, P. and Liu, X. and Worden, K.},
	year = {2020},
	pages = {106550},
}
%












\end{document}